\RequirePackage[svgnames,table]{xcolor}

\documentclass[10pt,letterpaper,twocolumn]{template/style}

\usepackage{graphicx}
\usepackage{float,epstopdf}
\usepackage{bbm}

\usepackage{microtype}

\usepackage[numbers]{natbib}
\setcitestyle{square}

\usepackage{subcaption}
\usepackage{booktabs}

\usepackage{amsmath}
\usepackage{amssymb}
\usepackage{mathtools}
\usepackage{amsthm}
\usepackage{dsfont}
\usepackage{makecell}
\usepackage{multirow} 
\usepackage{amsfonts} 
\usepackage{mathrsfs}
\usepackage[amssymb, thickqspace]{SIunits}
\DeclareFontFamily{U}{rsfs}{\skewchar\font127}
\DeclareFontShape{U}{rsfs}{m}{n}{%
  <5> <5.5> <6> rsfs5
  <7> rsfs7
  <8> <9> <10> <10.95> <12> <14.4> <17.28> <20.74> <24.88> rsfs10
}{}
\DeclareFontShape{OML}{eur}{m}{n}{%
  <5> <5.5> <6> <7> <8> <9> gen * eurm
  <10> <10.95> <12> <14.4> <17.28> <20.74> <24.88> eurm10
}{}
\usepackage{enumitem}
\usepackage{pgfplotstable}
\pgfplotsset{compat=1.18}
\usepackage{lipsum}		

\usepackage{microtype}
\usepackage{graphicx}
\usepackage{booktabs} 
\usepackage[table]{xcolor}
\usepackage{arydshln}
\usepackage[normalem]{ulem} 

\usepackage{cases}
\usepackage{wrapfig}

\usepackage{url}

\usepackage{thmtools}
\usepackage{thm-restate}
\usepackage{tabu}

\definecolor{huskypurple}{HTML}{4B2E83}

\usepackage{titletoc}

\usepackage{listings}
\lstdefinestyle{promptstyle}{
  basicstyle=\ttfamily\footnotesize,
  breaklines=true,
  breakautoindent=false,
  breakindent=0pt,
  postbreak=\mbox{\textcolor{gray}{$\hookrightarrow$}\space},
  columns=fullflexible,
  keepspaces=true,
  frame=single,
  framesep=5pt,
  xleftmargin=6pt,
  xrightmargin=6pt,
  aboveskip=8pt,
  belowskip=8pt,
  showstringspaces=false,
}

\usepackage{fontawesome5}   
\makeatletter
\def\munderbar#1{\underline{\sbox\tw@{$#1$}\dp\tw@\z@\box\tw@}}
\makeatother

\newtheorem{definition}{Definition}[section]

\newtheorem{corollary}[definition]{Corollary}

\AddToHook{cmd/appendix/before}{%
  \setcounter{axiom}{0}%
}

\newcommand{\be}{\begin{equation}}
\newcommand{\ee}{\end{equation}}
\newcommand{\bea}{\begin{equation*}\begin{aligned}}
\newcommand{\eea}{\end{aligned}\end{equation*}}

\newcommand{\monobold}[1]{{\fontfamily{lmtt}\bfseries\selectfont #1}}

\newtcolorbox{simpleElegantQuote}{
    colback=AliceBlue!50!White,
    colframe=RoyalBlue!75!Black,
    boxrule=0.5pt,
    arc=2mm,
    boxsep=4pt,
    left=10pt, right=10pt,
    top=8pt, bottom=8pt,
    fontupper=\itshape,
}

\title{Memory-Efficient Training-Free Acceleration of Diffusion Transformers with BaryCache}

\runningtitle{Memory-Efficient Training-Free Acceleration of Diffusion Transformers with BaryCache}


\newcommand{\affmark}[1]{%
  \textsuperscript{%
    {\usefont{T1}{pbk}{m}{n}\textcolor{BgPrimary600}{\textbf{#1}}}%
  }%
}

\author{Chengjie Lu\affmark{1}, Tianchi Deng\affmark{2}, Zhengqi He\affmark{1}, Zhijian Gao\affmark{1}, Huisi Wu\affmark{3}, Xueliang Li\affmark{2}\affmark{†}\\
\affmark{1}College of Electronics and Information Engineering, Shenzhen University\\
\affmark{2}School of Artificial Intelligence, Shenzhen University\\
\affmark{3}College of Computer Science and Software Engineering, Shenzhen University\\
{\footnotesize\affmark{†}Corresponding author}\\
\faGithub~\textbf{Source Code:} \href{https://github.com/Kiramei/BaryCache}{\monobold{https://github.com/Kiramei/BaryCache}}
}

\begin{document}
\raggedbottom

\vspace{-1mm}
\begin{abstract}
    \section*{Abstract}
    Diffusion Transformers achieve high-fidelity image and video generation, but their iterative sampling remains expensive, for each denoising step requires large matrix operations. Existing cache-based acceleration reduces redundant computation yet increases the VRAM footprint by storing intermediate states, which can directly constrain inference batch size. In this work, we propose a training-free acceleration method that performs stepwise forecasting for DiT sampling using a Barycentric Extrapolator. By leveraging barycentric extrapolation, our predictor is numerically stable and alleviates oscillatory artifacts analogous to the Runge phenomenon during forward forecasting. Across extensive experiments on both image and video generation, our approach provides a favorable trade-off between memory usage and perceptual quality, while delivering up to 3.30x end-to-end sampling speedup compared with baseline DiT inference.
\end{abstract}

\maketitle

\section{Introduction}
\label{sec:introduction}
Denoising diffusion probabilistic models (DDPMs)~\cite{hoDDPM} have become a dominant paradigm for generative modeling, often surpassing generative adversarial networks~\cite{park2019SPADE, brock2019biggan, Goodfellowgan} in synthesizing high-fidelity images at scale. Recent work increasingly incorporates Transformer architectures into diffusion backbones to better capture long-range dependencies, culminating in Diffusion Transformers (DiTs). DiTs have achieved strong performance on class-to-image (C2I)~\cite{peebles2023dit}, text-to-image (T2I)~\cite{chen2023pixartalpha, chen2024pixartsigma}, and text-to-video (T2V)~\cite{kong2024hunyuanvideo} tasks. However, scaling model capacity and resolution exacerbates the tension between generation quality and computational cost~\cite{liang2024scalinglawsdiffusiontransformers}.

To mitigate this trade-off, a broad set of acceleration strategies has been explored. Quantization~\cite{li2023qdiffusion} and VAE-based training optimizations~\cite{yao2025vavae} reduce memory and training overhead, but they often require retraining or finetuning. Sampling-side approaches improve efficiency by reducing iteration counts via deterministic trajectories~\cite{songDDIM} or higher-order solvers~\cite{lu2022dpm}. Flow-based models~\cite{lauRealNVP, kingma2018glow, liu2022flow} enable exact likelihood estimation, yet they can struggle with memory efficiency in high-dimensional settings~\cite{helminger2021lossy}.

\begin{figure}[t]
  \begin{center}
    \centerline{\includegraphics[width=0.95\columnwidth]{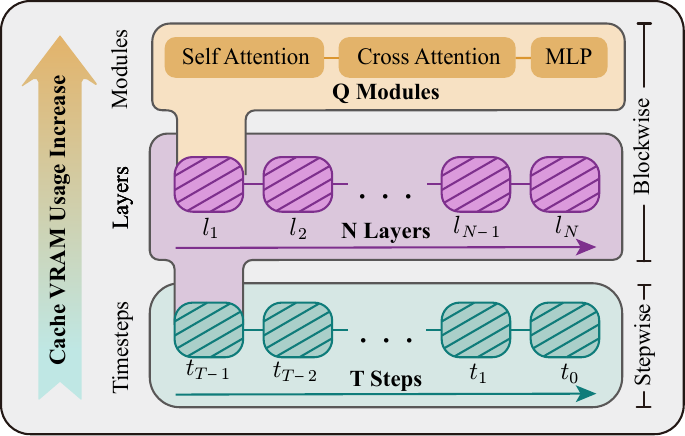}}
    \caption{
      Caching targets for different strategies. Reusing features across diffusion steps is \emph{stepwise} caching, while reusing features within a step across layers/modules is \emph{blockwise} caching. For $N$ diffusion steps, blockwise caching may require storing up to $N \times Q$ intermediate tensors compared to stepwise caching, which can significantly constrain batch size and parallelism.
    }
    \label{figure:intro}
  \end{center}
  \vskip -0.2in
\end{figure}

Among training-free approaches, caching methods reduce repeated computation by reusing intermediate activations. In the \emph{stepwise} caching line, DeepCache~\cite{ma2024deepcache} exploits temporal redundancy by reusing high-level features across adjacent diffusion steps. $\Delta$-DiT~\cite{chen2024delta-dit} further shows that diffusion inference can be separated into stages, where early and late DiT blocks play different roles. TeaCache~\cite{teacache} leverages time-embedding signals to build a caching-and-skipping schedule. While these methods can provide meaningful speedups, their quality often degrades substantially at high acceleration ratios. In parallel, \emph{blockwise} strategies operate at finer granularity within a DiT step. FORA~\cite{selvaraju2024fora} recycles module features to mitigate step errors; token-level reuse underlies ToCa~\cite{toca}, and the more aggressive DuCa~\cite{zou2025rethinkingtokenwisefeaturecaching} attains higher speedups by aggressive skipping. TaylorSeer~\cite{TaylorSeer2025} moves beyond reuse to forecasting, improving perceptual quality under acceleration. However, as illustrated in Figure~\ref{figure:intro}, blockwise caching can incur substantial inference-time memory overhead, which limits parallelization.

\begin{figure*}
    \centering
    \includegraphics[width=\textwidth]{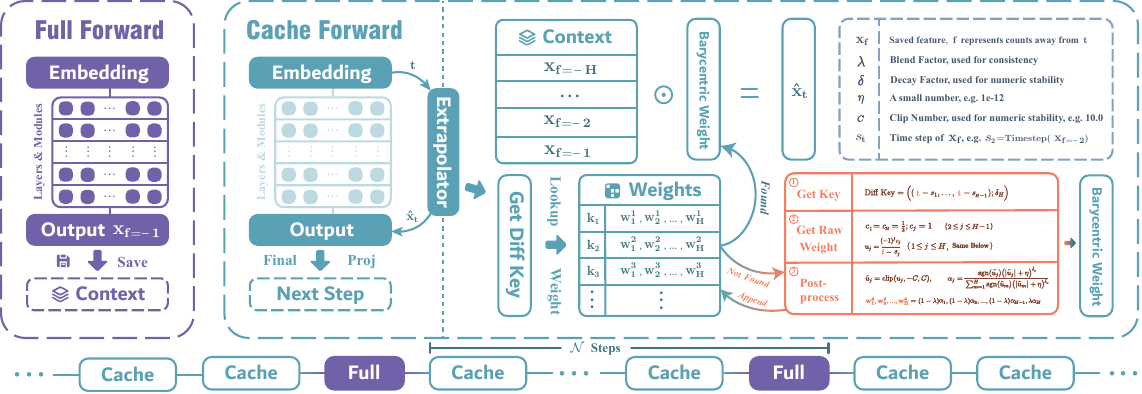}
    \caption{Overview of BaryCache. At each diffusion step, the model either performs a full forward pass or a cache forward pass, where the Barycentric Extrapolator yields weights and predicts the current stepwise feature from a cache context of recently computed full-step outputs. Weight post-processing stabilizes the barycentric combination for extrapolation. BaryCache balances memory usage and fidelity via stepwise feature prediction from prior steps under equal-spaced schedules.}
    \label{fig:main_method_overview}
    \vskip -0.2in
\end{figure*}

To better balance memory usage and generation quality, we propose \textbf{BaryCache}, a training-free acceleration method that predicts future \emph{stepwise} features in DiTs using our Barycentric Extrapolator. Originally, barycentric interpolation~\cite{berrutBarycentric} is an efficient reformulation of polynomial interpolation closely connected to Chebyshev interpolation~\cite{floydcheb, Trefethen2019approximation}, reducing computational overhead relative to the standard Lagrange form. In contrast, local derivative-based extrapolation can suffer from numerical instabilities such as Runge oscillations~\cite{GrasselliPelinovsky2008}, which become pronounced for longer-range prediction. Our Barycentric Extrapolator leverages barycentric interpolation theory to construct prediction polynomials for stepwise feature forecasting. We observe that standard barycentric weights can yield irregular combinations under certain input schedules; accordingly, we design a more stable weight combination inspired by trapezoidal endpoint coefficients, and apply additional weight post-processing to further improve stability. Moreover, we note that the barycentric weights depend only on the input schedule, specifically, the step distances between the cached context and the current step, and are independent of DiT features. This enables an \emph{online decoupling} of the extrapolator into (i) weight construction and (ii) weight application, where weights can be precomputed and stored for reuse. Extensive qualitative and quantitative results on three representative pretrained DiT-based models including the C2I model DiT/XL-2~\cite{peebles2023dit}, the T2I model PixArt-\(\Sigma\)~\cite{chen2024pixartsigma} and the T2V model Hunyuan~\cite{kong2024hunyuanvideo}, demonstrate that BaryCache achieves a favorable balance between memory footprint and generation quality, reaching up to \(3.30\times\) speedup under diverse settings.

Our contributions are summarized as follows:
\begin{itemize}
    \item We propose \textbf{BaryCache}, a training-free acceleration method that predicts \emph{stepwise} DiT features to better balance memory usage and generation quality.
    \item We develop \textbf{Barycentric Extrapolator} for \emph{stepwise} feature prediction, and adopt a stabilized weight construction inspired by trapezoidal endpoint coefficients to mitigate Runge-oscillation-like instabilities during extrapolation.
    \item We introduce an \textbf{online decoupling} technique that separates weight construction from application, enabled by the observation that barycentric weights depend only on the input schedule, allowing for precomputation and reuse.
    \item We validate the effectiveness of BaryCache on C2I, T2I, and T2V generation, achieving up to \(3.30\times\) speedup while maintaining high perceptual quality.
\end{itemize}

\section{Related Works}

\subsection{Diffusion Model}
Diffusion models~\cite{hoDDPM} have become a cornerstone of modern generative modeling, achieving state-of-the-art performance in high-fidelity image synthesis~\cite{peebles2023dit, chen2023pixartalpha, chen2024pixartsigma} and video generation~\cite{kong2024hunyuanvideo, wan2025}. Early diffusion architectures based on U-Nets~\cite{hoDDPM, rombach2021highresolution} were highly effective for iterative denoising, but scaling to higher resolutions and richer conditioning signals exposed limitations in long-range modeling capacity and computational efficiency. Transformer-based diffusion backbones, exemplified by the DiT~\cite{peebles2023dit}, address these challenges by leveraging the expressiveness and scalability of Transformers, enabling improved generation quality and flexibility. Subsequent work has further enhanced DiT with techniques such as self-conditioning~\cite{jiang2025sra, zhao2024dynamic, zhao2025dyditdynamicdiffusiontransformers}, extending its applicability to more complex and multimodal generation settings. Large-scale video generation results~\cite{kong2024hunyuanvideo} further highlight the promise of DiT-style architectures for modeling long-range temporal structure and real-world dynamics.

\subsection{Sampling Step Reduction}
Reducing sampling steps while maintaining generation quality is a long-standing goal in diffusion models. Early DDPMs~\cite{hoDDPM} require hundreds of iterations, motivating step-reduction methods such as DDIM~\cite{songDDIM}, which introduces deterministic non-Markovian sampling. The DPM-Solver family~\cite{lu2022dpm, lu2022dpm++, zheng2023dpm} further accelerates inference by modeling the reverse process as an ODE and applying higher-order solvers. Other directions include flow-based methods such as Rectified Flow~\cite{liu2022flow} and distillation-based approaches~\cite{salimans2022progressive, luhman2021knowledge}. More recently, Consistency Models~\cite{song2023consistency} and their variants~\cite{issenhuth2024improving, lee2024truncated} achieve strong few-step generation, while theoretical results~\cite{yangimproved} indicate that multi-step updates improve stability. Together, these works establish sampling step reduction as a core strategy for accelerating diffusion models.

\subsection{Cache Acceleration}
Another major direction for training-free acceleration is to cache and reuse intermediate activations, exploiting temporal redundancy across neighboring diffusion steps. A line of work focuses on \emph{stepwise} caching. DeepCache~\cite{ma2024deepcache} shows that high-level U-Net features evolve smoothly over time and can be reused using a simple non-uniform caching policy with limited quality degradation. In the DiT setting, $\Delta$-DiT~\cite{chen2024delta-dit} proposes a structure-aware strategy that accelerates blocks asymmetrically based on their functional roles during sampling. TeaCache~\cite{teacache} leverages timestep-embedding signals and polynomial corrections to guide caching and skipping decisions, demonstrating strong gains especially for video generation.

In parallel, \emph{blockwise} caching targets redundancy within a single diffusion step by reusing features across layers, modules, or tokens. FORA~\cite{selvaraju2024fora} accelerates DiTs by recycling redundant attention and MLP activations without modifying model parameters. Token-level methods such as ToCa~\cite{toca} and TokenCache~\cite{lou2024tokencachingdiffusiontransformer} further refine granularity by caching subsets of tokens selected by attention statistics or learned predictors, often combined with layer-wise ratio schedules to control error accumulation. DuCa~\cite{zou2025rethinkingtokenwisefeaturecaching} adopts a more aggressive tokenwise strategy and achieves higher speedups by skipping additional computation. Beyond reuse, TaylorSeer~\cite{TaylorSeer2025} advances to \emph{prediction}-based acceleration by forecasting future features via Taylor-series extrapolation, improving perceptual quality under aggressive acceleration. Compared with blockwise approaches, these methods may incur substantial memory overhead due to storing many intermediate tensors. Therefore, we develop BaryCache that implements stepwise prediction mechanisms, while aiming to retain high fidelity while keeping inference-time memory manageable.

\section{Method}

\subsection{Preliminary}
\label{sec:preliminary}
\textbf{Diffusion Models.} Diffusion models are a class of generative probabilistic models formulated through two stochastic processes: a forward process that incrementally perturbs data with noise, and a reverse process that learns to reconstruct data by denoising. Given a sample \( \mathbf{x}_0 \sim q(\mathbf{x}_0) \), the forward process produces a sequence \( \{\mathbf{x}_t\}_{t=1}^T \) over \( T \) timesteps as
\begin{equation}
\mathbf{x}_t =
\sqrt{\alpha_t} \, \mathbf{x}_{t-1} +
\sqrt{1-\alpha_t} \, \boldsymbol{\epsilon}_t,
\end{equation}
where \( \boldsymbol{\epsilon}_t \sim \mathcal{N}(\mathbf{0}, \mathbf{I}) \), and \( \{\alpha_t\} \) denotes a predefined noise schedule. The reverse process approximates the intractable posterior \( q(\mathbf{x}_{t-1}|\mathbf{x}_t) \) via a parameterized Gaussian transition:

\begin{figure}[t]
  \begin{center}
    \centerline{\includegraphics[width=0.90\columnwidth]{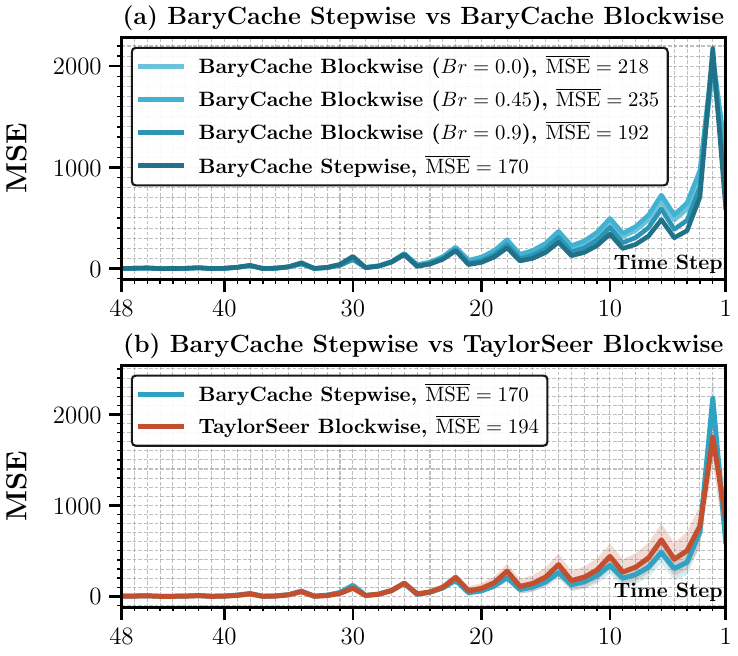}}
    \caption{
      Mean square error comparison in the preliminary evaluation. (a) shows that stepwise prediction performs better than blockwise variants. (b) shows that the stepwise version of BaryCache surpasses TaylorSeer in later steps, despite a larger error at step 2, indicating that BaryCache preserves quality while requiring less memory.
    }
    \label{figure:method01}
  \end{center}
  \vskip -0.25in 
\end{figure}

\begin{equation}
p_\theta(\mathbf{x}_{t-1}|\mathbf{x}_t) =
\mathcal{N}\big(
\mathbf{x}_{t-1};
\boldsymbol{\mu}_\theta(\mathbf{x}_t, t), \, \beta_t \mathbf{I}
\big),
\end{equation}
where the mean is computed by a neural network \( \boldsymbol{\epsilon}_\theta \) as
\begin{equation}
\boldsymbol{\mu}_\theta(\mathbf{x}_t, t) =
\frac{1}{\sqrt{\alpha_t}} \left(
\mathbf{x}_t -
\frac{1-\alpha_t}{\sqrt{1-\bar{\alpha}_t}} \,
\boldsymbol{\epsilon}_\theta(\mathbf{x}_t, t)
\right),
\end{equation}
with \( \bar{\alpha}_t = \prod_{i=1}^t \alpha_i \). The model parameters \( \theta \) are optimized to minimize a variational bound or a score-matching loss.

\textbf{Stepwise Features in DiTs.} 
Building on the probabilistic foundation above, a DiT realizes the reverse process as a composition of $T$ denoising steps. At timestep $t$, the model takes a noisy input $\mathbf{x}_t^{in}$, a conditioning input $y$, and a timestep embedding as inputs. The noisy input $\mathbf{x}_t^{in}$ is embedded into token latents $\mathbf{x}_t$, while the conditioning and timestep inputs are embedded and fused during the forward pass. A DiT contains $N$ layers, denoted by $\mathcal{F}^l$ for layer $l$. Since each layer contains $Q$ modules, denoted by $M_q$ for module $q$, the denoising function of a layer can be refined as $\mathcal{F}^l_q$ and expressed as:

\begin{equation}
\label{eq:blockwise_feature}
\mathcal{F}^l_q(u) = 
\begin{cases}
u + M_1^l(u), 
& q=1, \\
u + M_q^l(\mathcal{F}^l_{q-1}(u))
& 2\le q\le Q.
\end{cases}
\end{equation}

The stepwise feature $x_t'$ is then obtained as:

\begin{equation}
    \label{eq:stepwise_feature}
    \mathbf{x}_t' = (F_Q^L \circ F_Q^{L-1} \circ \cdots \circ F_Q^{2} \circ F_Q^{1})(\mathbf{x}_t).
\end{equation}

Equations~\ref{eq:blockwise_feature} and~\ref{eq:stepwise_feature} show that modules and layers contribute to stepwise features through incremental changes. Prior work~\cite{selvaraju2024fora, TaylorSeer2025} has observed that blockwise feature trajectories evolve smoothly across diffusion steps. This motivates the following corollary.

\begin{corollary}
  \label{cor:stepwise_feature}
  The stepwise features $\mathbf{x}_t'$ evolve smoothly across diffusion steps and are therefore predictable.
\end{corollary}

\begin{table*}[t]
\caption{\centering Quantitative comparison in text-to-image generation.}
\label{table:pixart_sigma_performance}
\vskip -0.1in
\centering
\setlength\tabcolsep{7pt}
\resizebox{0.98\textwidth}{!}{
\begin{tabular}{l | c c c c | c c}
\toprule
\textbf{Method}
&\multicolumn{4}{c|}{\textbf{Acceleration \& Runtime}}
& \textbf{Image} $\uparrow$ 
& {\bf Clip} $\uparrow$ \\
{\bf PixArt-$\mathbf{\Sigma}$}~\cite{chen2024pixartsigma}
& \textbf{FLOPs (T)}$\downarrow$ 
& \textbf{Latency (s)}$\downarrow$ 
& \textbf{Speedup} $\uparrow$ 
& \textbf{VRAM (GB)} $\downarrow$ 
& {\bf Reward}
& {\bf Score} \\
\midrule
{\textbf{$\text{[Vanilla]: DDIM-50 steps}$}}                                                                         & 450.52 & 4.974 & 1.00$\times$ & 3.363 & 0.7258 & 30.89 \\
\midrule
{\textbf{$\text{[Vanilla]: DDIM-15 steps}$}}                                                                        & 135.16 & 1.484 & 3.35$\times$ & 3.363  & 0.6577 & 31.04 \\
{\textbf{{$\Delta$}-DiT ($\mathcal{N}=4$)}~\cite{chen2024delta-dit}}                                                & 126.15 & 1.651 & 3.01$\times$ & 3.651  & 0.6246 & 30.96 \\
{\textbf{FORA ($\mathcal{N}=4$)}~\cite{selvaraju2024fora} \textcolor{red}{\textdagger}}                             & 125.68 & 3.939 & 1.26$\times$ & 19.255 & 0.5024 & 30.79 \\
{\textbf{ToCa ($\mathcal{N}=4$)}~\cite{toca} \textcolor{red}{\textdagger}}                                          & 189.08 & 4.957 & 1.00$\times$ & 27.453 & 0.5679 & 30.83 \\
{\textbf{DuCa ($\mathcal{N}=4$)}~\cite{zou2025rethinkingtokenwisefeaturecaching} \textcolor{red}{\textdagger}}      & 167.94 & 4.593 & 1.08$\times$ & 27.453 & 0.5842 & 30.87 \\
{\textbf{TeaCache ($l=0.65$)}~\cite{teacache}}                                                                      & 126.15 & 1.666 & 2.98$\times$ & 3.795  & 0.6478 & 31.01 \\
{\textbf{TaylorSeer ($\mathcal{N}=4$)}~\cite{TaylorSeer2025} \textcolor{red}{\textdagger}}                          & 125.88 & 2.874 & 1.73$\times$ & 21.913 & 0.7097 & 30.98 \\
\rowcolor{gray!20}
{\textbf{BaryCache ($\mathcal{N}=4, H=2$)}}                                                                              & 126.15 & 1.863 & 2.67$\times$ & 3.941  & {\bf 0.7220} & {\bf 30.92} \\
\midrule
{\textbf{$\text{[Vanilla]: DDIM-11 steps}$}}                                                                        & 99.11  & 1.088 & 4.57$\times$ & 3.363  & 0.5844 & 31.05 \\
{\textbf{{$\Delta$}-DiT ($N=6$)}~\cite{chen2024delta-dit}}                                                          & 90.10  & 1.284 & 3.87$\times$ & 3.651  & 0.5001 & 31.04 \\
{\textbf{FORA ($\mathcal{N}=6$)}~\cite{selvaraju2024fora}\textcolor{red}{\textdagger}}                              & 89.77  & 3.153 & 1.58$\times$ & 19.255 & 0.2723 & 30.54 \\
{\textbf{ToCa ($\mathcal{N}=6$)}~\cite{toca} \textcolor{red}{\textdagger}}                                          & 160.21 & 4.240 & 1.17$\times$ & 27.453 & 0.3541 & 30.92 \\
{\textbf{DuCa ($\mathcal{N}=6$)}~\cite{zou2025rethinkingtokenwisefeaturecaching} \textcolor{red}{\textdagger}}      & 132.04 & 3.522 & 1.41$\times$ & 27.453 & 0.3642 & 30.86 \\
{\textbf{TeaCache ($l=0.95$)}~\cite{teacache}}                                                                      & 99.11  & 1.394 & 3.57$\times$ & 3.795  & 0.6130 & 31.17 \\
{\textbf{TaylorSeer ($\mathcal{N}=6$)}~\cite{TaylorSeer2025} \textcolor{red}{\textdagger}}                          & 89.98  & 2.772 & 1.79$\times$ & 21.913 & 0.6537 & 30.92 \\
\rowcolor{gray!20}
{\textbf{BaryCache ($\mathcal{N}=6, H=2$)}}                                                                              & 90.10  & 1.505 & 3.30$\times$ & 3.941  & {\bf 0.6654} & {\bf 31.10} \\
\bottomrule
\end{tabular}}
\vspace{2pt}\
\par
\raggedright
\footnotesize
\quad \textcolor{red}{$\dagger$} Block-wise caching improves quality but incurs VRAM overhead, limiting the batch size.
Same in Table~\ref{table:hunyuanvideo_performance} and Table~\ref{table:dit_xl2_performance}.
\vskip -0.15in
\end{table*}

\begin{table*}[t]
\vskip -0.12in
\caption{\centering Quantitative comparison in text-to-video generation.}
\label{table:hunyuanvideo_performance}
\vskip -0.12in
\centering
\setlength\tabcolsep{7pt}
\resizebox{0.98\textwidth}{!}{
\begin{tabular}{l | c c c | c c c}
\toprule
\textbf{Method}
&\multicolumn{3}{c|}{\textbf{Acceleration \& Runtime}}
&\multicolumn{3}{c}{\textbf{Generation Quality}}
\\
{\bf HunyuanVideo}~\cite{kong2024hunyuanvideo}
& \textbf{Latency (s)}$\downarrow$ 
& \textbf{Speedup} $\uparrow$ 
& \textbf{VRAM (GB)} $\downarrow$ 
& {\bf PSNR}$\uparrow$
& {\bf SSIM}$\uparrow$
& {\bf LPIPS}$\downarrow$ \\
\midrule
{\textbf{$\text{[Vanilla]: Euler-50 steps}$}}                                                                        & 231.240 & 1.000$\times$ & 31.171 & -       & -      & -      \\
\midrule
{\textbf{ToCa ($\mathcal{N}=5$)}~\cite{toca} \textcolor{red}{\textdagger}}                                     & 87.580  & 2.640$\times$ & 46.521 & 15.8272 & 0.5069 & 0.4437 \\
{\textbf{DuCa ($\mathcal{N}=5$)}~\cite{zou2025rethinkingtokenwisefeaturecaching} \textcolor{red}{\textdagger}} & 82.480  & 2.804$\times$ & 46.515 & 15.8641 & 0.5122 & 0.4440 \\
{\textbf{TeaCache ($l=0.4$)}~\cite{teacache}}                                                                  & 84.920  & 2.723$\times$ & 31.525 & 17.0962 & 0.5984 & 0.3775 \\
{\textbf{TaylorSeer ($\mathcal{N}=5$)}~\cite{TaylorSeer2025} \textcolor{red}{\textdagger}}                     & 87.940  & 2.630$\times$ & 60.771 & 15.9317 & 0.5214 & 0.4269 \\
\rowcolor{gray!20}
{\textbf{BaryCache ($\mathcal{N}=5, H=2$)}}                                                                         & 84.650  & 2.732$\times$ & {\bf 30.805} & {\bf 19.0159} & {\bf 0.6829} & {\bf 0.2992} \\
\midrule
{\textbf{ToCa ($\mathcal{N}=7$)}~\cite{toca} \textcolor{red}{\textdagger}}                                     & 81.290  & 2.845$\times$ & 46.521 & 15.3712 & 0.5045 & 0.4962 \\
{\textbf{DuCa ($\mathcal{N}=7$)}~\cite{zou2025rethinkingtokenwisefeaturecaching} \textcolor{red}{\textdagger}} & 75.570  & 3.060$\times$ & 46.515 & 15.3682 & 0.5058 & 0.4980 \\
{\textbf{TeaCache ($l=0.5$)}~\cite{teacache}}                                                                  & 77.060  & 3.001$\times$ & 31.525 & 17.3969 & 0.5921 & 0.3806 \\
{\textbf{TaylorSeer ($\mathcal{N}=7$)}~\cite{TaylorSeer2025} \textcolor{red}{\textdagger}}                     & 75.190  & 3.075$\times$ & 60.771 & 15.4674 & 0.5104 & 0.4716 \\
\rowcolor{gray!20}
{\textbf{BaryCache ($\mathcal{N}=7, H=2$)}}                                                                         & 76.640  & 3.017$\times$ & {\bf 30.805} & {\bf 17.9988} & {\bf 0.6496} & {\bf 0.3463} \\
\bottomrule
\end{tabular}}
\vskip -0.1in
\end{table*}

\begin{table*}[t]
\vskip 0.12in
\caption{\centering Quantitative comparison in class-to-image generation for DiT-XL/2.}
\label{table:dit_xl2_performance}
\vskip -0.12in
\centering
\setlength\tabcolsep{7pt}
\resizebox{0.98\textwidth}{!}{
\begin{tabular}{l | c c c c | c c | c}
\toprule
\textbf{Method}
&\multicolumn{4}{c|}{\textbf{Acceleration \& Runtime}}
&\multicolumn{3}{c}{\textbf{Generation Quality}} \\
{\bf DiT-XL/2}~\cite{peebles2023dit}
& \textbf{FLOPs (T)}$\downarrow$ 
& \textbf{Latency (s)}$\downarrow$ 
& \textbf{Speedup} $\uparrow$ 
& \textbf{VRAM (GB)} $\downarrow$ 
& \textbf{FID} $\downarrow$ 
& \textbf{sFID} $\downarrow$ 
& \textbf{IS {$\uparrow$}} \\
\midrule
{\textbf{$\text{[Vanilla]: DDIM-50 steps}$}}                                                                        & 23.74 & 0.349 & 1.00$\times$ & 6.037 & 2.27 & 4.34  & 239.61 \\
\midrule
{\textbf{$\text{[Vanilla]: DDIM-25 steps}$}}                                                                        & 11.87 & 0.234 & 1.49$\times$ & 6.037 & 2.91 & 4.54  & 229.97 \\
{\textbf{{$\Delta$}-DiT ($\mathcal{N}=3$)}~\cite{chen2024delta-dit}}                                                & 8.07  & 0.182 & 1.91$\times$ & 6.037 & 3.58 & \underline{4.66}  & 217.19 \\
{\textbf{FORA ($\mathcal{N}=5$)}~\cite{selvaraju2024fora} \textcolor{red}{\textdagger}}                             & 5.24  & 0.190 & 1.84$\times$ & 6.467 & 5.81 & 9.74  & 200.64 \\
{\textbf{ToCa ($\mathcal{N}=5$)}~\cite{toca} \textcolor{red}{\textdagger}}                                          & 7.72  & 0.303 & 1.15$\times$ & 6.569 & 4.41 & 6.21  & 213.46 \\
{\textbf{DuCa ($\mathcal{N}=5$)}~\cite{zou2025rethinkingtokenwisefeaturecaching} \textcolor{red}{\textdagger}}      & 6.04  & 0.232 & 1.50$\times$ & 6.569 & 4.81 & 6.75  & 207.04 \\
{\textbf{TeaCache ($l=0.2$)}~\cite{teacache}}                                                                       & 8.55  & 0.197 & 1.77$\times$ & 6.057 & 3.38 & 4.73  & 218.06 \\
{\textbf{TaylorSeer ($\mathcal{N}=5$)}~\cite{TaylorSeer2025} \textcolor{red}{\textdagger}}                          & 5.24  & 0.215 & 1.62$\times$ & 8.483 & \textbf{2.73} & 5.44  & \textbf{233.35} \\
\rowcolor{gray!20}           
{\textbf{BaryCache ($\mathcal{N}=5, H=2$)}}                                                                         & 5.24  & 0.181 & 1.93$\times$ & {\bf 5.529} & \underline{3.27} & \textbf{4.61}  & \underline{219.67} \\
\midrule             
{\textbf{$\text{[Vanilla]: DDIM-20 steps}$}}                                                                        & 5.70  & 0.194 & 1.80$\times$ & 6.037 & 3.49 & 4.88  & 224.42 \\
{\textbf{{$\Delta$}-DiT ($N=4$)}~\cite{chen2024delta-dit}}                                                          & 7.75  & 0.151 & 2.31$\times$ & 6.037 & 5.02 & \underline{5.33}  & 201.71 \\
{\textbf{FORA ($\mathcal{N}=6$)}~\cite{selvaraju2024fora}\textcolor{red}{\textdagger}}                              & 4.77  & 0.164 & 2.12$\times$ & 6.467 & 9.20 & 14.97 & 171.28 \\
{\textbf{ToCa ($\mathcal{N}=6$)}~\cite{toca} \textcolor{red}{\textdagger}}                                          & 7.25  & 0.297 & 1.17$\times$ & 6.569 & 5.77 & 7.54  & 200.86 \\
{\textbf{DuCa ($\mathcal{N}=6$)}~\cite{zou2025rethinkingtokenwisefeaturecaching}\textcolor{red}{\textdagger}}       & 5.85  & 0.202 & 1.73$\times$ & 6.569 & 5.90 & 7.58  & 198.84 \\
{\textbf{TeaCache ($l=0.4$)}~\cite{teacache}}                                                                       & 5.22  & 0.140 & 2.49$\times$ & 6.057 & 6.71 & 6.62  & 185.89 \\
{\textbf{TaylorSeer ($\mathcal{N}=6$)}~\cite{TaylorSeer2025} \textcolor{red}{\textdagger}}                          & 4.77  & 0.202 & 1.73$\times$ & 8.483 & \textbf{3.17} & 6.47  & \textbf{225.77} \\
\rowcolor{gray!20}           
{\textbf{BaryCache ($\mathcal{N}=6, H=2$)}}                                                                         & 4.77  & 0.160 & 2.18$\times$ & {\bf 5.529} & \underline{3.91} & \textbf{5.27}  & \underline{209.77} \\
\bottomrule
\end{tabular}}
\vskip -0.1in
\end{table*}


\subsection{Barycentric Extrapolator}
Following Corollary~\ref{cor:stepwise_feature}, we introduce the Barycentric Extrapolator, the core predictor in BaryCache, highlighted in yellow in Figure~\ref{fig:main_method_overview}.

\textbf{Cache context.} The Barycentric Extrapolator maintains a dynamic cache $\mathcal{H}_t$ at the current timestep $t$. This cache stores a history of feature maps from previous evaluations. Let $H$ denote the history length, measured by the relative distance from $t$. The cache is defined as a sequence of feature tensors:

\begin{equation}
  \mathcal{H}_t = \{ \mathbf{x}_{f} \mid f \in \{-\text{H}, -\text{H}+1, \dots, -1\} \}
\end{equation}

Here, the index $f$ represents the relative temporal distance from the current step $t$. Specifically, $\mathbf{x}_{f=-1}$ is the feature map from the most recent saved cache, and $\mathbf{x}_{f=-H}$ is the oldest feature in the cache context. 
In the stepwise scheduling paradigm, this cache acts as a sliding window. At each full step, the oldest feature $\mathbf{x}_{f=-H}$ is evicted, indices are shifted ($\mathbf{x}_{i} \leftarrow \mathbf{x}_{i+1}$), and the newly computed feature is inserted at $\mathbf{x}_{f=-1}$.

\textbf{Get Key.} Instead of extrapolating raw feature magnitudes directly, the Barycentric Extrapolator operates on schedules. Because weight computation is independent of DiT features, we decouple weight computation from feature aggregation. A \textit{Diff Key} is introduced for weight lookup, avoiding repeated weight calculation. Let $s_1, \cdots, s_{H} \in S_t$ denote the timesteps corresponding to the cached features $ \mathbf{x}_{f=-\mathbf{1}},  \cdots,  \mathbf{x}_{f=-\mathbf{H}}$, where $t$ is the input timestep. The \textit{Diff Key} is defined as a tuple containing relative timestep differences and a decay factor $\delta_H$:

\begin{equation}
  \text{k}_i = \left( (t - s_{1}, \dots, t - s_{H-1}); \delta_H \right), 
\end{equation}

where $i$ represents the unique index of a key.

\textbf{Get Raw Weight.} The calculation of weights $w_j$ is central to the method's stability. The coefficients are assigned based on a trapezoidal-like rule for the history buffer of size $H$:

\begin{equation}  
  c_1 = c_H = 0.5, c_j = 1 \quad (2 \le j \le H-1)
\end{equation}

This assignment is consistent with the weights of the composite trapezoidal rule for numerical integration. In the context of barycentric extrapolation on equispaced nodes, weights proportional to $(-1)^j c_j$, where $c_j$ are trapezoidal weights, correspond to the \textit{Floater-Hormann rational interpolant} with degree $d=0$ or $d=1$. This specific choice is known to be stable and pole-free.

The raw barycentric weight $u_j$ for the $j$-th node in the history relative to the target time $t$ involves the inverse distance kernel. As $s_j$ is the timestep of the cached feature $\mathbf{x}_{f=-j}$, the raw weight is computed as:

\begin{equation}
  u_j = \frac{c_j (-1)^j}{t - s_j},
\end{equation}

where the factor $(-1)^j$ is implicit in standard barycentric formulations to handle the sign alternation of the nodal polynomial denominator.

\textbf{Postprocess.} In order to prevent numerical overflow when $t$ is dangerously close to a history node $t_j$, which would cause the denominator $t - s_j$ to vanish, the weights are clipped by a stability constant $C$:

\begin{equation}
  \tilde{u}_j = \text{clip}(u_j, -C, C) \\
            = \max(-C, \min(C, u_j)),
\end{equation}

where a value of $C=10$ is suggested in Barycentric Extrapolator. This ensures that no single node dominates the interpolation sum excessively due to proximity singularities.

To further enhance stability, a decay factor $\delta \in (0, 1)$ is introduced. This factor modulates the influence of long-range history extrapolations. Because $\delta$ is closely related to the context length $H$, we empirically set $\delta_H = 1.0$ for $H=2$, $\delta_H = 0.95$ for $H=3$, and $\delta_H = 0.7$ for $H=4$. The final weight $w_j$ is computed as:

\begin{equation}
  \alpha _j = \frac{\text{sgn}(\tilde{u}_j)( | \tilde{u}_j | + \eta) ^{\delta_H}}{\sum_{m=1}^H \text{sgn}(\tilde{u}_m) (| \tilde{u}_m | + \eta) ^{\delta_H}},
\end{equation}

where $\text{sgn}(\cdot)$ is the sign function. Finally, a blend factor $\lambda \in [0, 1]$ is applied to mix the extrapolated value with the most recent cached feature $\mathbf{x}_{f=-1}$. Adjusting $\lambda$ controls the consistency of the prediction. The final weight vector is given by:

\begin{equation}
\begin{aligned}
  \mathbf{W_i} 
  &= [\mathbf{w_1^i}, \mathbf{w_2^i}, \cdots, \mathbf{w_H^i}] \\
  &= [(1-\lambda)\alpha_1, \cdots, (1-\lambda)\alpha_{H-1}, \lambda \alpha_H]
\end{aligned}
\end{equation}

Combining these elements, the Barycentric Extrapolator obtains the weight combination used for future predictions. We use a small constant $\eta$ (e.g., 1e-12) to keep the computation numerically well-defined.

\subsection{BaryCache}
Based on the Barycentric Extrapolator, we can formalize BaryCache, which is colored in peacock blue in Figure~\ref{fig:main_method_overview}. First, at each diffusion timestep $t$, BaryCache decides whether to perform a full computation or a cache forward. If a full computation is executed, the DiT processes the input normally, and the resulting feature $\mathbf{x}_t'$ is stored in the cache $\mathcal{H}_t$. If a cache forward is chosen, BaryCache invokes the Barycentric Extrapolator to predict the feature $\hat{\mathbf{x}}_t$ using the cached features in $\mathcal{H}_t$, as follows:
 
\begin{equation}
  \hat{\mathbf{x}}_t = \mathbf{W_i}\mathcal{H}^{\mathsf T},
\end{equation}

where $\mathbf{W_i}$ is the weight vector obtained from the Barycentric Extrapolator using the current timestep $t$. 

For preliminary validation, we sample a batch of images with the basic DiT framework and compare the mean square error of different caching methods, as shown in Figure~\ref{figure:method01}. The results demonstrate that stepwise prediction outperforms blockwise variants, and BaryCache surpasses TaylorSeer in later steps, indicating its effectiveness in mitigating Runge-phenomenon-like instability while reducing memory usage.

\section{Experiment}

\begin{figure*}[t]
    \centering
    \includegraphics[width=0.93\linewidth]{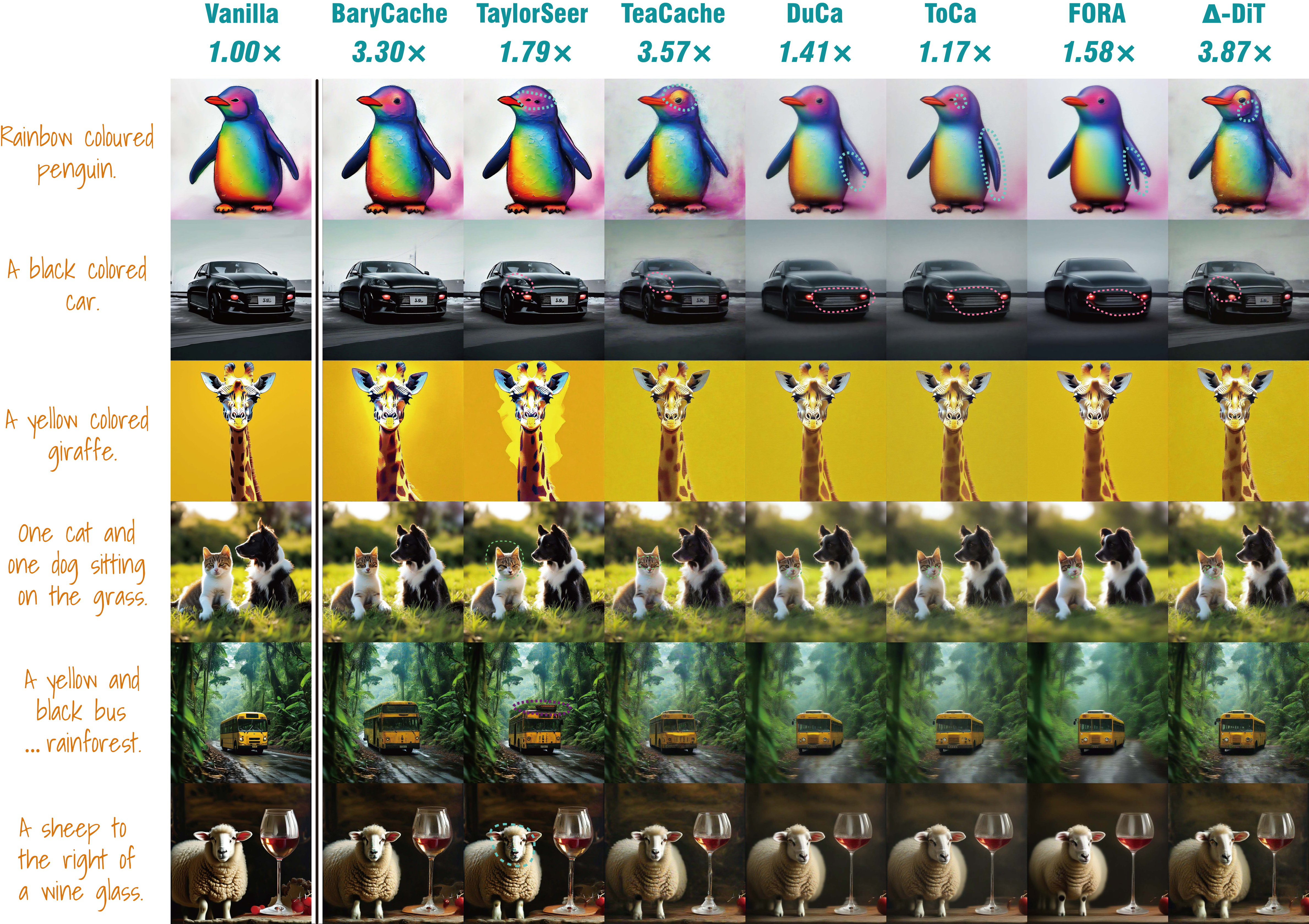}
    \caption{\centering Qualitative comparison on PixArt-$\Sigma$. }
    \label{fig:pixart_comparison}
    \vskip -0.2in

\end{figure*}

\begin{figure*}[t]
    \centering
    \includegraphics[width=0.93\linewidth]{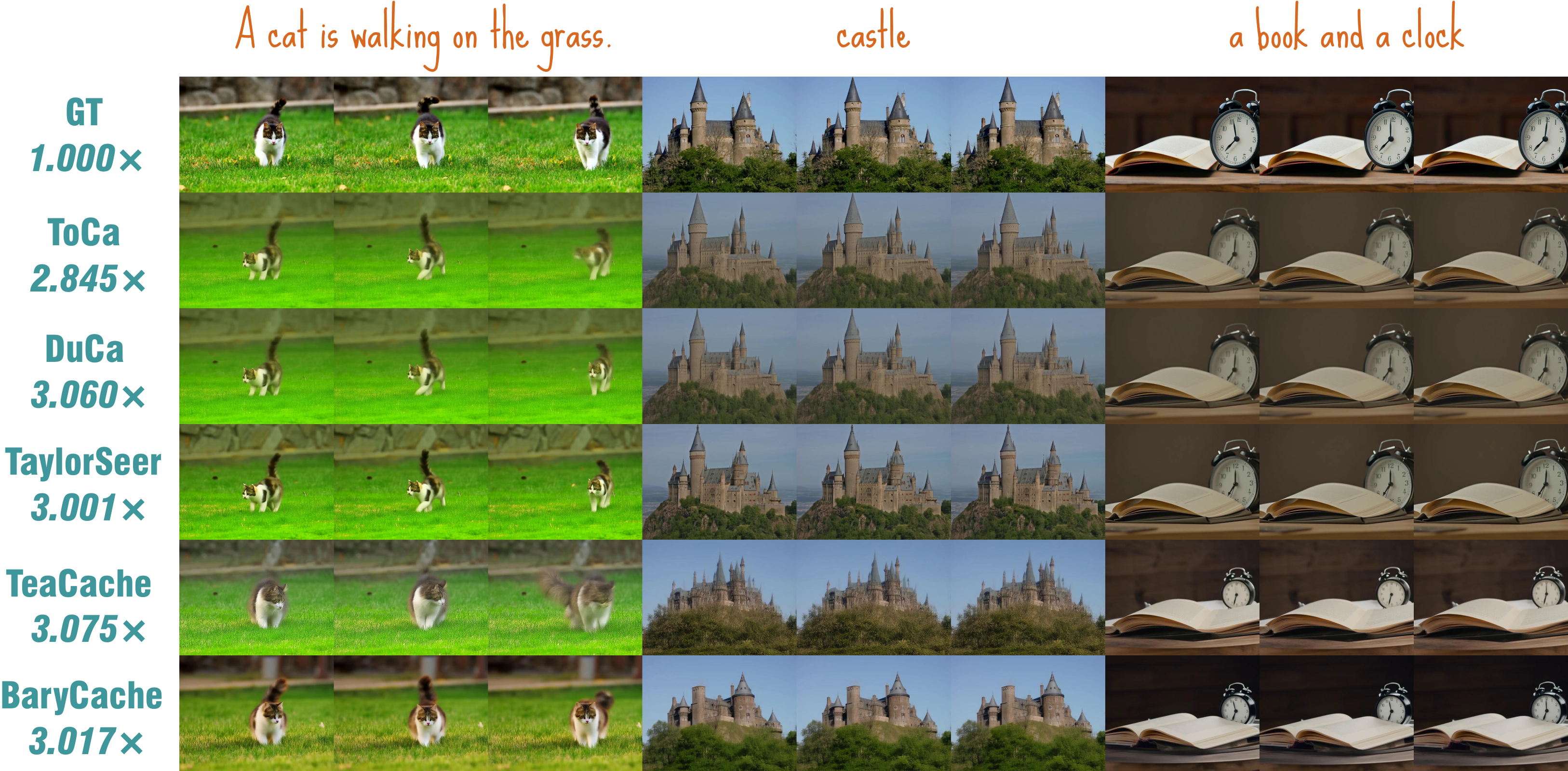}
    \caption{\centering Qualitative comparison on HunyuanVideo.}
    \label{fig:hunyuan_comparison}
    \vskip -0.15in
\end{figure*}

\begin{figure}[t]
    \vskip -0.05in
    \centering
    \includegraphics[width=0.93\linewidth]{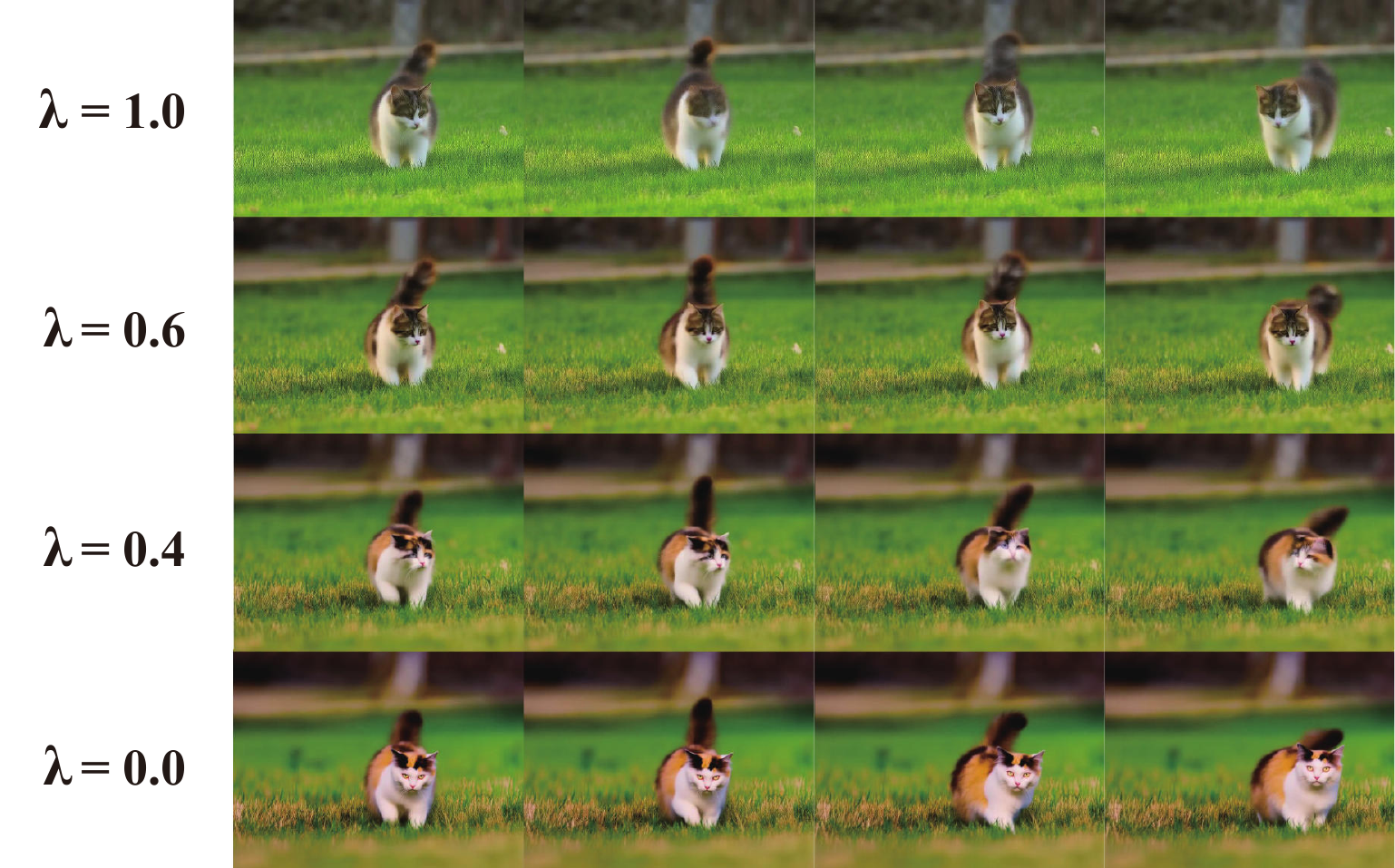} 
    \caption{Qualitative ablation of $\lambda$ on HunyuanVideo.}
    \label{fig:blendfactor1}
    \vskip -0.15in
\end{figure} 

\subsection{Experimental Setup}
We evaluate BaryCache on three representative DiT-based generators: DiT-XL/2~\cite{peebles2023dit} for C2I ($256\times256$), PixArt-\(\Sigma\)~\cite{chen2024pixartsigma} for T2I ($1024\times1024$), and HunyuanVideo~\cite{kong2024hunyuanvideo} for T2V ($640\times480$). Quantitative results are reported in Tables~\ref{table:dit_xl2_performance}, \ref{table:pixart_sigma_performance}, and \ref{table:hunyuanvideo_performance}, respectively. We use official pretrained weights and default inference configurations with 50 sampling steps (DDIM for C2I/T2I and Euler for T2V). For C2I, we evaluate FID-50K~\cite{fid50k}, sFID~\cite{nashsfid}, and Inception Score (IS)~\cite{salimans2016improved}. For T2I, we evaluate all 200 DrawBench~\cite{saharia2022photorealistic} prompts using CLIPScore~\cite{clipscore} and ImageReward~\cite{imagereward}. For T2V, we randomly select 100 prompts from EvalCrafter~\cite{liu2023evalcrafter} and report PSNR, SSIM, and LPIPS~\cite{zhang2018unreasonableeffectivenessdeepfeatures} against the baseline outputs.

C2I and T2I experiments are conducted on NVIDIA RTX 5090 GPUs. For T2V, VRAM and instance analyses are conducted on NVIDIA A800 GPUs, while PSNR/SSIM/LPIPS evaluation is conducted on H100 GPUs. In T2I, we decouple text embedding from sampling to avoid recomputing text features for identical prompts; therefore, T2I VRAM measurements focus on the DiT step-forward process, whereas C2I/T2V measurements include embedding-model memory. We use ImageReward 1.0 from the official implementation, CLIPScore with CLIP-ViT-B/32 via TorchMetrics, and the Guided Diffusion implementation for FID, sFID, and IS. Unless otherwise specified, the random seed is 2026, the C2I CFG scale is 1.5, and batch sizes are 4 for C2I/T2I and 1 for T2V. We set $\lambda$ to zero for C2I and T2I, while for T2V we use $\lambda = 1.0$ and $\lambda = 0.8$ under $\mathcal{N}=5$ and $\mathcal{N}=7$, respectively.

\subsection{Quantitative Results}
Results on C2I, T2I, and T2V indicate that BaryCache provides a favorable trade-off between memory footprint and generation quality. On T2I (Table~\ref{table:pixart_sigma_performance}), stepwise baselines degrade substantially at higher skip ratios. For instance, under $\mathcal{N}=6$, $\Delta$-DiT attains ImageReward $0.5001$ and CLIPScore $31.04$, whereas BaryCache improves both metrics to $0.6654$ and $31.10$, while remaining within a modest runtime gap (only $0.5\times$ slower than the fastest caching method under the same setting). In contrast, blockwise caching methods such as FORA, ToCa, DuCa, and TaylorSeer require substantially larger activation buffers---often approaching an order of magnitude more VRAM than the original model---which limits batch size and can increase latency. Moreover, blockwise approaches may accumulate prediction errors across modules, and in some settings yield weaker overall quality than stepwise methods despite their higher memory cost. On T2V, BaryCache achieves the best consistency among caching baselines under a severe skip ratio of seven, reaching PSNR $17.9988$, SSIM $0.6496$, and LPIPS $0.3463$ (Table~\ref{table:hunyuanvideo_performance}). Importantly, these gains are obtained with low VRAM usage, improving memory efficiency. On C2I, although TaylorSeer achieves stronger FID/IS, BaryCache reduces inference memory by nearly \textbf{3GB}, which materially improves parallelization and practical throughput. Compared with the remaining baselines, BaryCache consistently delivers higher quality while maintaining competitive speed (e.g., $2.18\times$ for BaryCache versus $2.49\times$ for TeaCache at $\mathcal{N}=6$).

\subsection{Qualitative Results}
Figures~\ref{fig:pixart_comparison} and~\ref{fig:hunyuan_comparison} provide qualitative evidence that BaryCache better preserves semantic structure and fine details under aggressive acceleration. On PixArt-\(\Sigma\) (Figure~\ref{fig:pixart_comparison}), for the prompt \textit{A yellow colored giraffe}, most baselines fail to maintain the intended coloration, while TaylorSeer introduces an unexpected background. For \textit{One cat and one dog sitting on the grass.}, several methods distort the cat's facial structure, whereas BaryCache produces a coherent composition. For the \textit{black car} example, many methods exhibit severe global blurring, and even TaylorSeer blurs localized regions (e.g., the right headlight), while BaryCache retains sharper structure. On HunyuanVideo (Figure~\ref{fig:hunyuan_comparison}), BaryCache yields the highest visual consistency relative to the baseline across diverse prompts. For \textit{A cat is walking on the grass} and \textit{castle}, BaryCache preserves salient spatial details and motion dynamics: in the former, competing methods often produce backward or unstable stepping patterns, whereas BaryCache generates a natural forward gait. For \textit{a book and a clock}, BaryCache better preserves textual details, correctly rendering the clock numerals, while other methods disrupt fine-grained text structure.

\subsection{Ablation on History Length}
In our main experiments, we use a cache history of two full-computed steps to maximize efficiency. We further ablate the history length $H$ under stepwise and eq-spaced mode, while we report results in Table~\ref{table:ablation_n_h}, where we keep $\lambda$ zero and adopt $\delta = 1.0, 0.95, 0.7$ for $\mathcal{N}=2,3,4$, respectively. Although we adjust the decay ratio to maintain numerical stability as $H$ increases, the best ImageReward is achieved at $H=2$, indicating that the most efficient configuration is also the most effective in terms of perceptual quality.

\begin{table}[ht]
\centering
\caption{\centering Ablation study on cache history length.}
\label{table:ablation_n_h}
\vskip -0.1in
\begin{small}
\begin{tabular}{llcc}
\toprule
\multicolumn{2}{l}{\textbf{Ablation}} & \textbf{Image Reward}$\uparrow$ & \textbf{Clip Score}$\downarrow$ \\
\midrule
\multirow{3}{*}{$\mathcal{N}=4$} & $\mathit{H}=2$ & \textbf{0.7220} & 30.92 \\
                                 & $\mathit{H}=3$ & 0.6936 &  \textbf{30.99} \\
                                 & $\mathit{H}=4$ & 0.6914 & 30.94 \\
\midrule
\multirow{3}{*}{$\mathcal{N}=5$} & $\mathit{H}=2$ & \textbf{0.6812} & \textbf{30.94} \\
                                 & $\mathit{H}=3$ & 0.6557 & 30.85 \\
                                 & $\mathit{H}=4$ & 0.6510 & 30.93 \\
\midrule
\multirow{3}{*}{$\mathcal{N}=6$} & $\mathit{H}=2$ & \textbf{0.6654} & 31.10 \\
                                 & $\mathit{H}=3$ & 0.6285 & \textbf{31.16} \\
                                 & $\mathit{H}=4$ & 0.5958 & 31.06 \\
\bottomrule
\end{tabular}
\end{small}
\end{table}

\subsection{Ablation on Schedules and Coefficients}
\begin{table}[ht]
\centering
\caption{\centering Ablations on different modes and schedules.}
\label{tab:ablation_mode_schedule}
\vskip -0.1in
\begin{small}
\begin{tabular}{llcrr}
\toprule
{\bf Mode} & {\bf Schedule} & {\bf Setting} & {\bf IR$\uparrow$} & {\bf CS$\uparrow$} \\
\midrule
\multirow{6}{*}{\shortstack{\textbf{Stepwise}\\$\lambda=0$}}
 & {\bf Eq-spaced} & $N=4$ & 0.7220 & 30.92 \\
 & {\bf Adaptive } & $\ell=0.13$ & 0.6594 & 30.83 \\
\cmidrule(lr){2-5}
 & {\bf Eq-spaced} & $N=5$ & 0.6812 & 30.94 \\
 & {\bf Adaptive } & $\ell=0.16$ & 0.6034 & 30.84 \\
 \cmidrule(lr){2-5}
 & {\bf Eq-spaced} & $N=6$ & 0.6654 & 31.10 \\
 & {\bf Adaptive } & $\ell=0.2$  & 0.5881 & 30.97 \\
\midrule
\multirow{6}{*}{\shortstack{\textbf{Blockwise} \\ $\lambda=0$ \\ $\mathbf{Br}=0$}}
 & {\bf Eq-spaced} & $N=4$ & 0.7088 & 30.96 \\
 & {\bf Adaptive } & $\ell=0.13$ & 0.6494 & 30.80 \\
 \cmidrule(lr){2-5}
 & {\bf Eq-spaced} & $N=5$ & 0.6445 & 30.87 \\
 & {\bf Adaptive } & $\ell=0.16$ & 0.5918 & 30.81 \\
 \cmidrule(lr){2-5}
 & {\bf Eq-spaced} & $N=6$ & 0.6579 & 31.11 \\
 & {\bf Adaptive } & $\ell=0.2$  & 0.5685 & 30.88 \\
\bottomrule
\end{tabular}
\end{small}
\end{table}

\begin{table}[ht]
\centering
\caption{\centering  Ablation study on $\lambda$ and $\textbf{Br}$.}
\label{tab:ablation_bf_cr_merged}
\vskip -0.1in
\begin{small}
\begin{tabular}{cccc}
\toprule
{\bf Ablation} & {\bf Value} & {\bf Image Reward} $\uparrow$ & {\bf Clip Score} $\uparrow$ \\
\midrule
\multirow{6}{*}{\shortstack{$\lambda$ \\ (\emph{stepwise}, \\ $\mathcal{N}=4$, \\ $H=2$ )}}
& 0.0 & \textbf{0.7220} & 30.92 \\
& 0.2 & 0.7002 & 30.95 \\
& 0.4 & 0.6857 & 30.95 \\
& 0.6 & 0.6887 & 31.00 \\
& 0.8 & 0.6566 & 31.00 \\
& 1.0 & 0.6200 & \textbf{31.01} \\
\midrule
\multirow{6}{*}{\shortstack{$\textbf{Br}$ \\ (\emph{blockwise}, \\ $\lambda = 0$, \\ $\mathcal{N}=4$, \\ $H=2$)}}
& 0.0 & 0.7088 & 30.96 \\
& 0.2 & 0.6888 & 30.95 \\
& 0.4 & 0.6965 & 30.96 \\
& 0.6 & 0.6991 & 30.97 \\
& 0.8 & 0.6935 & \textbf{30.98} \\
& 1.0 & \textbf{0.7181} & 30.92 \\
\bottomrule
\end{tabular}
\end{small}
\vskip -0.1in
\end{table}

\begin{figure}[t]
    \vskip 0.05in
    \centering
    \includegraphics[width=0.92\linewidth]{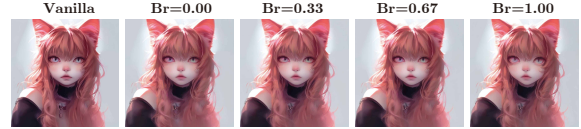}
    \caption{\centering Qualitative ablation of the blockwise ratio \textbf{Br}.}
    \label{fig:blockwise_ratio_ablation}
    \vskip -0.15in
\end{figure}

\begin{figure}[t]
    \centering
    \includegraphics[width=0.94\linewidth]{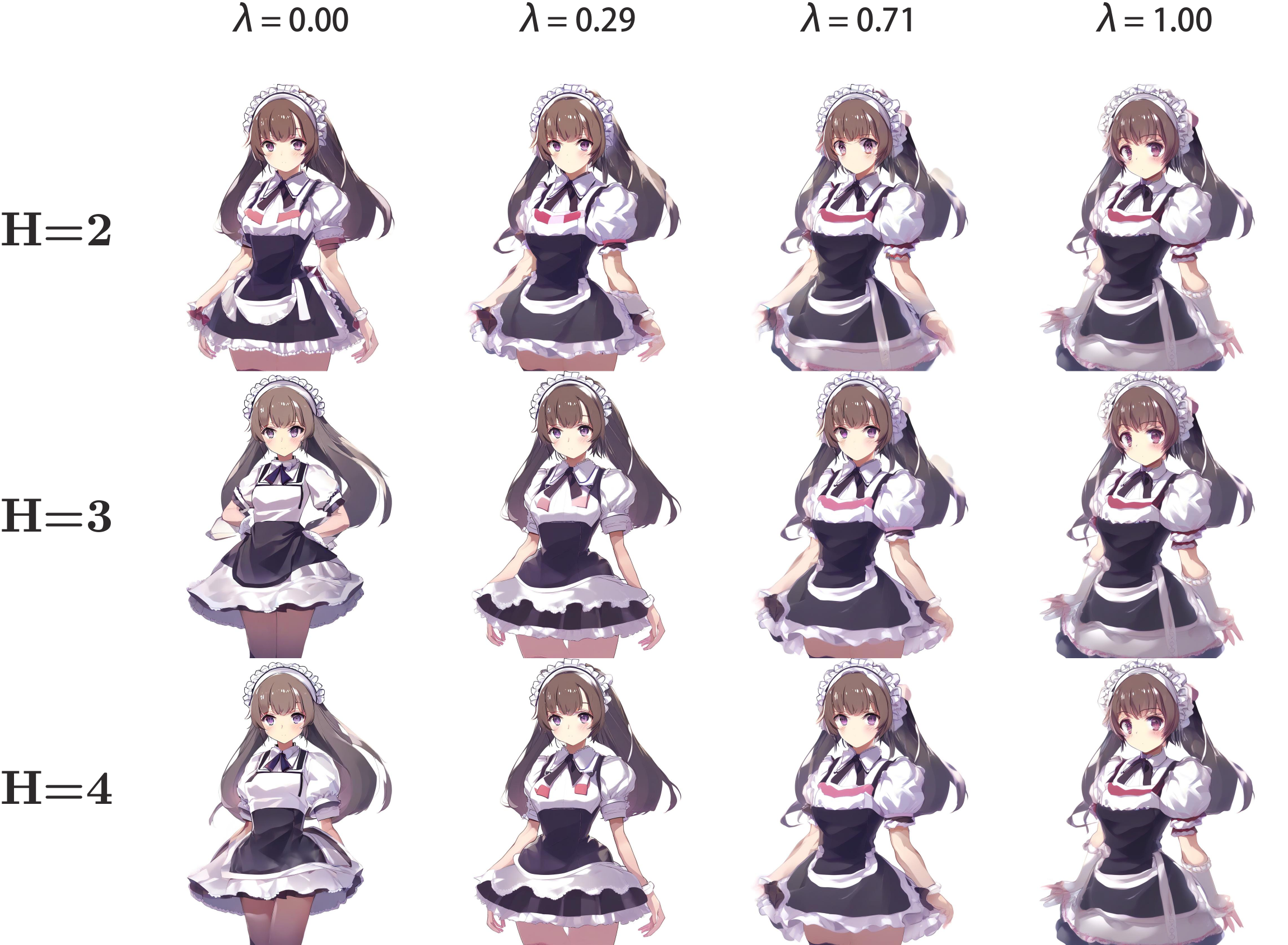}
    \caption{\centering Qualitative ablation of $\lambda$ on PixArt-$\Sigma$.}
    \label{fig:blendfactor2}
\end{figure}

\begin{figure}[t]
    \centering
    \includegraphics[width=0.94\linewidth]{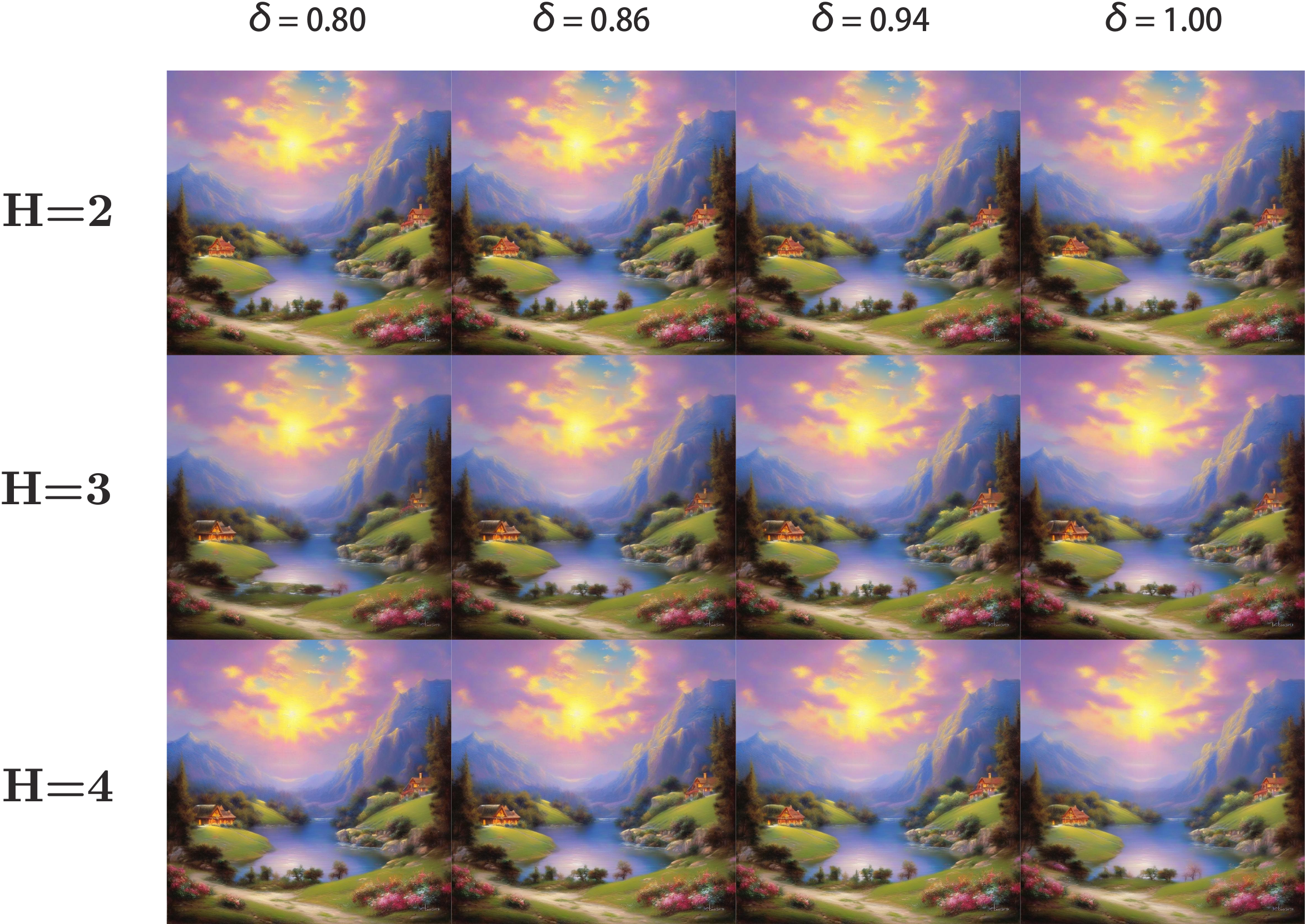}
    \caption{\centering Qualitative ablation of $\delta$ on PixArt-$\Sigma$.}
    \label{fig:decay1}
    \vskip -0.1in
\end{figure}

We further evaluate BaryCache under different scheduling and coefficient choices. Table~\ref{tab:ablation_mode_schedule} compares equal-spaced and adaptive schedules under stepwise and blockwise modes. The adaptive schedule follows a TeaCache-style rule that accumulates relative feature-probe changes and triggers full computation when the accumulated drift exceeds a threshold. Under matched full-computation budgets, equal-spaced schedules consistently outperform adaptive ones, suggesting that regular anchor placement is more compatible with our barycentric extrapolator.

Table~\ref{tab:ablation_bf_cr_merged} studies the blend factor $\lambda$ in stepwise mode and the blockwise ratio $Br$ in blockwise mode. Increasing $\lambda$ provides a controllable interpolation between the barycentric estimate and the most recent cached feature, leading to smooth changes in ImageReward and CLIPScore. For blockwise mode, $Br=1.0$ performs best among the tested settings, which is equivalent to stepwise caching. Figure~\ref{fig:blockwise_ratio_ablation} further visualizes this trend: as $\mathbf{Br}$ increases, the generated structure remains visually close to the vanilla result while retaining the low-memory stepwise cache design. These results support using the default stepwise configuration for the best quality-memory trade-off. We also visualize the effects of $\lambda$ and $\delta$ on PixArt-$\Sigma$ in Figures~\ref{fig:blendfactor2} and~\ref{fig:decay1}, respectively, while a T2V qualitative result for the $\lambda$ ablation is shown in Figure~\ref{fig:blendfactor1}. In these cases, we observe that larger $\lambda$ enables more aggressive feature forecasting and can alter the generated structure, while $\delta$ controls fine details, such as the trees in Figure~\ref{fig:decay1}. These results suggest that $\lambda$ and $\delta$ can be tuned to balance fidelity and efficiency.

\section{Discussion}

During the development of BaryCache, we observed that the method can occasionally produce images or videos with unexpected patterns, such as the sheep beside the wine glass in Figure~\ref{fig:pixart_comparison} and the forest details in the \textit{castle} example in Figure~\ref{fig:hunyuan_comparison}. We estimate that these artifacts may originate from local object-information loss during skipped forward passes, which the current extrapolator cannot fully correct. In addition, the decay factors in BaryCache are empirically selected, and their sensitivity across different models and tasks has not been systematically studied. Future work will focus on improving the extrapolator and exploring more adaptive decay-factor strategies to further improve the robustness of BaryCache.
\section{Conclusion}
In this work, we introduced BaryCache, a stepwise feature extrapolation method for accelerating DiT inference. By leveraging the Barycentric Extrapolator, BaryCache predicts current stepwise activations with high fidelity while maintaining manageable memory usage. Extensive experiments on DiT-based image and video generation tasks demonstrate that BaryCache achieves a strong balance between quality and memory usage, outperforming existing caching and step-reduction methods on most tasks. Overall, BaryCache offers a practical and memory-efficient solution for accelerating diffusion models, paving the way for real-time applications in high-quality generative modeling. Future work may explore more adaptive extrapolation techniques for further improving generation quality.

\section*{Acknowledgment}
This work was supported by the Guangdong Provincial Natural Science Foundation -- General Program (Grant No. 2025A1515011568); the Project of Shenzhen Science and Technology Innovation Bureau -- General Project (Grant No. JCYJ20250604182252068); the Guangdong Provincial Department of Education Key Areas Special Project for University Scientific Research (Grant No. 2024ZDZX1015); and the Internal Fund of National Engineering Laboratory for Big Data System Computing Technology (Grant No. SZU-BDSC-IF2024-05).

{\raggedright
\bibliography{main}
\bibliographystyle{template/abbrvnat}
}

\end{document}